%% file: main.tex
\pdfoutput=1
\documentclass[10pt,twocolumn,letterpaper]{article}

\usepackage{graphicx}
\usepackage{amsmath}
\usepackage{amssymb}
\usepackage{booktabs}
\usepackage[table,xcdraw,dvipsnames]{xcolor}

\usepackage{multirow}
\usepackage[table,xcdraw]{xcolor}

\usepackage[pagebackref,breaklinks,colorlinks]{hyperref}
\usepackage[capitalize]{cleveref}
\crefname{section}{Sec.}{Secs.}
\Crefname{section}{Section}{Sections}
\Crefname{table}{Table}{Tables}
\crefname{table}{Tab.}{Tabs.}

\newcommand{\ourmethodtable}{{CornerFormer(Ours)}\xspace}

\newcommand{\first}[1]{\textcolor{cyan}{#1}}
\newcommand{\second}[1]{\textcolor{orange}{#1}}

\input{math_definition}
\input{table_definition}

\begin{document}

\title{CornerFormer: Boosting Corner Representation for Fine-Grained Structured
Reconstruction }

\author{
Hongbo Tian\textsuperscript{1, 2} \thanks{This work was done when the authors were visiting Beike as interns.} 
\and Yulong Li\textsuperscript{2*}   
\and  Linzhi Huang\textsuperscript{1, 2} 
\and  Xu Ling\textsuperscript{1} 
 \and Yue Yang\textsuperscript{2} 
 \and Jiani Hu\textsuperscript{1 \thanks{Corresponding author.}}  \\
 {\textsuperscript{1}Beijing University of Posts and Telecommunications, \textsuperscript{2}Beike} \\
 {\tt\small\{tianhongbo,huanglinzhi,lingxu,jnhu\}@bupt.edu.cn} \\
 {\tt\small\{liyulong008, yangyue092\}@ke.com}
}
%
 

\maketitle

\begin{abstract}
Structured reconstruction is a non-trivial dense prediction problem, which extracts structural information (\eg, building corners and edges) from a raster image, then reconstructs it to a 2D planar graph accordingly. Compared with common segmentation or detection problems, it significantly relays on the capability that leveraging holistic geometric information for structural reasoning.
Current transformer-based approaches tackle this challenging problem in a two-stage manner, which detect corners in the first model and classify the proposed edges (corner-pairs) in the second model. 
However, they separate two-stage into different models and only share the backbone encoder.
Unlike the existing modeling strategies, we present an enhanced corner representation method: 1) It fuses knowledge between the corner detection and edge prediction by sharing feature in different granularity; 2) Corner candidates are proposed in four heatmap channels w.r.t its direction.
Both qualitative and quantitative evaluations demonstrate that our proposed method can better reconstruct fine-grained structures, such as adjacent corners and tiny edges. Consequently, it outperforms the state-of-the-art model by +1.9\%@F-1 on Corner and +3.0\%@F-1 on Edge.
\end{abstract}

\section{Introduction}
\label{sec:intro}

Structured reconstruction\cite{birchfield1999multiway,gallup2010piecewise,furukawa2009manhattan} is a fundamental task for rendering, effects mapping,  human-computer interaction, etc. 
It requires a thorough understanding of the geometrical information in the image and then represents it in vector form, such as CAD format.

In recent years, computer vision-based structured reconstruction has proven successful for the reconstruction of interior floor plans \cite{MonteFloor},\cite{FloorSP}.  
 While the outdoor structured reconstruction is still challenging, owing to its sophisticated structure. 
 It suffers from dense targets (\eg, redundant non-target structures, adjacent corners or tiny edges in Fig. \ref{fig:head}), which may lead to the collapse of the whole reconstruction.
Human vision goes beyond computers in structured reconstruction not only in the understanding of the overall structure but also in the local fine-grained structure.
Inspired by the strengths of human perceptual capability, we propose an improvement method introduced briefly in Fig. \ref{fig:head} for the current advanced work.
\input{texts/figures/head_img.tex}

At first, the reasoning process of humans on structure information requires global and local coordination. During perceiving corners, we have a basic understanding of high-level semantics, like edges and regions. 
Therefore, Corner detection and edge prediction are inseparable in such a reasoning process \cite{LETR},\cite{HAWP}. 
However, the current advanced methods ignore such phenomenon \cite{ConvMPN},\cite{HEAT}. 
They treat corner detection and edge prediction as two detachable modules, which can be employed alone.
It results in a severe loss of local information.
This is inconsistent with human cognitive processes.

Our method leverages corner detection to assist edge prediction by treating the feature proposal as the anchor of query embedding\cite{zhou2022centerformer}. 
Specifically, the fine-grained structural semantics is learned from local fine features according to the proposed corners, and the high-level geometric construction is learned from global coarse features\cite{HEAT}.

The other problem in the current methods is the quantization error\cite{Sampling-Argmax} caused by corner detection \cite{HEAT,zhou2019end, Faster-Rcnn}.
Adjacent corners at a very close distance will be mixed up after the non-maximum suppression(NMS).
In response to this problem, we designed the direction corner module to learn the corner according to its direction, which greatly alleviated the problem of adjacent corners and boosted corner representation.

Our method effectively compensates for the information loss between corner detection and edge prediction and is more sensitive to fine structures.
We evaluate the proposed method in outdoor architecture reconstruction ~\cite{IP} from satellite images and it performs well especially in the corner and edge scores. 
The contributions of our method can be summarised as follows:
\begin{itemize}
  \item [1)] 
  Fine-grained image features based on the corner proposal are used to assist edge prediction without the increase of parameters.
  Besides, a proposal feature enhancement module (PFEM) is designed to facilitate the convergence of fine features.
  \item [2)]
  A direction corner module is adopted for tiny edges in structured reconstruction to solve adjacent corners problem and to boost corner representation.
  \item [3)]
  Our work is proved to outperform the current competing methods in outdoor building architecture with complex environments through quantitative and qualitative experiments.
\end{itemize}

\section{Related Work}
\label{sec:related}
\subsection{Structured Reconstruction}

We introduce structured reconstruction work based on two different structure inference processes. One is top-down method, and the other is bottom-up method. 
The top-down method is based on the fact that the structure is a simple 1D polygonal loop \cite{acuna2018efficient, castrejon2017annotating,cabral2014piecewise}. 
This method employs an instance segmentation algorithm such as Mask R-CNN~\cite{MaskR-CNN} for room proposal, and then iteratively optimizes the proposed rooms.
Specifically, FloorSP~\cite{FloorSP} and Cabral \etal~\cite{cabral2014piecewise} explore the shortest path algorithm to reconstruct an indoor floorplan. 
MonteFloor~\cite{MonteFloor} based on Monte Carlo Tree Search maximizes a suitable objective
function for planar graphs.
However, these methods do not perform well in complex outdoor environments with nested loops and the outdoor structured reconstruction mainly adopts the method of bottom-up ~\cite{IP,ppgnet,hedau2009recovering}.
IP~\cite{IP} and Exp-cls~\cite{expcls} based on heuristics extract structural information and infer the graph structure. 
They are powerful but require the design of rigorous rules for extensive optimization and search process.
ConvMPN ~\cite{ConvMPN} and HEAT ~\cite{HEAT} sought solutions without structural optimization by directly fusing line segment information into image features. 

Structure inference based on bottom-up relies heavily on the results of corner detection, while current advanced methods ~\cite{ConvMPN, HAWP} all design a detachable module to better focus on the improvement of piecewise results. 
Our model can combine the two-stage models to improve the two-stage performance synchronously.
\subsection{Vision Transformer}
Transformer~\cite{VisionTransformer} already has a capacity in computer vision that matches and even surpasses convolutional neural networks. 
It has the entire receptive field of a feature map. 
For the same reason, even the transformer model adopted by DETR ~\cite{DETR} shows strong performance in the object detection task, but the memory access sharply increases with the feature scale and the computational complexity grows quadratically with the spatial size. 
Deformable-DETR~\cite{DeformableDETR} adaptively learns the feature in limited locations from the feature maps and Feature Pyramid Networks(FPN) divides the features into different levels to alleviate the problem.
Both above can effectively reduce the computational burden of the model but are difficult to converge during training by using a dummy query~\cite{LETR, DETR}.
Improvement schemes ~\cite{liu2022dab,meng2021conditional,zhang2022dino} based on DETR turn out that the initialization of anchors makes the transformer model converge easily. 
This paper directly takes the anchor feature from the corner model as the input of the edge prediction and further introduces the feature information of corner detection.
By boosting the corner representation, our method accelerates the convergence of query embedding and performers well in optimizing corner proposal.
\subsection{Adjacent Corners Problem}
Both human detection and wireframe parsing tasks based on the bottom-up method need to detect corner primitives first. The traditional methods \cite{tompson2014joint,wei2016convolutional,yu2021lite} detect corners based on heatmap followed by NMS, but can not avoid quantization errors caused by NMS. 
Some classic work based on the regression  ~\cite{li2021human,luvizon20182d, Sampling-Argmax} solves the problem.
Wireframe parsing methods \cite{huang2018learning},~\cite{zhou2019end} alleviate this problem through patches of heatmap and regression. 
IP~\cite{IP} and DWP~\cite{dWP} perform a fine-grained design of corner bins according to the angle. 
Besides, R2V ~\cite{R2V} scatters different corners into different channels according to their types, which also avoids part of the problem.

Our direction corner detection is inspired by characteristics of corners ~\cite{morley2001ganet},~\cite{R2V}. 

In outdoor structure reconstruction, most of the adjacent corners will form tiny edges. We scatter the tiny edge endpoints into different channels according to the directions of the two endpoints of an edge to complete the detection of tiny edges.

\input{texts/figures/method.tex}
\section{Method}
\label{sec:method}
We have shown the complete network system in Fig. \ref{fig:method} and we will explain our model in six sections. 
1) The preliminaries in our method;
2) Corner detection with the corner decoder and direction corner proposal; 
3) Learnable corner feature extractor to extract corner features according to proposed corners; 
4) Proposal feature enhancement module;
5) Edge decoder for edge prediction; 
and 6) Loss function.

\subsection{Preliminaries}
\mypara{Pipeline.}
Corner detection is responsible for corner proposal and the proposed corners are grouped in pairs considered as candidate edges. 
During training, we take all GT edges as well as the wrong candidate edges (randomly combined based on the predicted corners and supplementary materials for details) to generate our candidate set. 
The maximum number of candidate edges in the candidate set is $T$, and all the  candidate edges in our set are delivered to the edge prediction for holistic prediction.
In the inference process, all the proposed corners $n$ are freely combined into $\bm{C^{2}_{n}}$ candidate edges for edge prediction. 
Finally, our model is evaluated based on the precision/recall/F-1 scores of the corner/edge/region primitives.

\mypara{Data preparation.}
The main notations go first to make our method more intuitive. 
The input is the  cropped satellite image $I\in\mathbb{R}^{H, W, 3}$. 
Extracted coarse features and fine features are represented as $F_{coarse}$ and $F_{fine}$ respectively.
For all the quantization results, we denote the proposal corner as $\bm{e}$, and the coordinate corresponding to the current corner is denoted as $(x,y)$. 
A candidate edge grouped by pair corners is denoted as $\bm{v}$.
A weight learned through a layer of linear mapping is denoted as $W$. 
For any one of the proposed corners, we extract corner feature $\Phi=\{\bm{\phi}_{l}\}_{l=0}^{L}$. 
$L$ denotes the total number of feature layers from corner decoder feature maps.
A set of candidate edge features is represented by $F_{v}=\{f_{v}^{t}\}_{t=1}^{T}$.
An individual edge proposal feature $f_{v}$ formed from coupled $\Phi$.

\mypara{Backbone.}
The backbone is based on Resnet~\cite{Resnet} and the stacked  convolutional layers from a multi-scale feature pyramid.
The 6-layers feature maps is corresponding to six different scales
$F^{l}\in\mathbb{R}^{\frac{H}{2^{l}},\frac{W}{2^{l}},C^{l}}$,$(l=0,1,\cdots,5)$.
Coarse-scale feature maps $l_{3}-l_{5}$ are assigned to the  encoder for extracting global coarse features. 
Fine-scale feature maps $l_{0}-l_{2}$ are assigned to the corner decoder for corner detection.
This is accord with HEAT~\cite{HEAT}.

\mypara{Image encoder.}
\label{sec:4.1}
To ensure the computation cost is acceptable, a multi-scale transformer encoder requires compressing large scale feature maps into a small size. 
Hence, the coarse-scale feature maps $l_{3}-l_{5}$ from the backbone are assigned for self-attention encoding. 
Both corner model and edge model employ the same encoder from deformable-DETR~\cite{DeformableDETR} for corner detection and edge prediction respectively. 
The key and query elements in transformer are pixels from the coarse-scale feature maps which are flattened to query embedding $\bm{q}\in\sum_{l=3}^{5}\frac{H \cdot W}{2^{2l}}, C$. 
At the same time, the central reference point and position encoding are provided for each feature map, which effectively reduces the computational complexity compared to DETR-style transformer encoder ~\cite{DETR}.

\subsection{Corner detection}
\input{texts/figures/driection_corner.tex}

\mypara{Corner decoder.} 
Based on the consensus that there exists only one corner within the 4×4 pixel patch, one layer of corner decoder was employed to decode all position embedding of 4×4 patches in an image. 
The output from the corner decoder is mixed with the backbone at fine-scale ones, $l_{0}-l_{2}$. 
This process is similar to matching the stacked hourglass network~\cite{newell2016stacked} with the transformer system. 
The output feature map from corner decoder is concatenated with the corresponding scale feature map from the backbone respectively by convolution and upsampling. 
We transform all the fine-scale feature maps to 128-dimensions for proposal feature extraction in Sec. \ref{sec:3.3}.

\mypara{Direction corner module.}
As we know, the two edge endpoints from an edge are in opposite directions and almost all adjacent corners from tiny edges in closed geometry.
The adjacent corners from a tiny edge can be detected by scattering into opposite direction channels natively. 

However, just two opposite direction channels are not sufficient.
If we detect all the corners based on the up and down directions,  adjacent corners in horizontal tiny edges can not be separated in the up-to-down corner channel.
So four directions are needed, and pairs of directions are opposite.

Specifically (See Fig. \ref{fig:direction_corner}), in the training process, we classify all corners into four categories according to the direction of current corner ray along the edge.
It brings corners with different semantic directions (up, down, left, right) and an corner from different edges will have multiple semantic directions at the same time.

Then, the loss function $L_{direct}$ is assembled to learn them onto four confidence maps $Y\in\mathbb{R}^{H, W, 4}$, using a Gaussian blur $G(x,y)$ as follows:
\begin{equation}
G(x,y)=\frac{1}{2\pi\sigma^{2}}exp(-\frac{(x-\check{x})^{2}+(y-\check{y})^{2}}{2\sigma^{2}})
\end{equation}
where $\check{x}$ and $\check{y}$  denote the corners with directional semantics and  $\sigma$ is the standard deviation depends on the gaussian kernel.
An additional line segment semantic segmentation loss $L_{seg}$ is used to supervise direction corner learning.
During inference, we perform non-maximum suppression (NMS) on the predicted confidence maps with different semantic corners, respectively. 
All the corners whose cluster centroid distance is less than L are clustered, where L=5.


\subsection{Learnable corner feature extractor}
\label{sec:3.3}
Using guidance like learnable position encoding effectively improves the decoding performance of transformer~\cite{HEAT}.
We further propose the local fine features from corner model according to the proposed corners.
The corner feature $\bm{\phi}^{l}(x,y)$ is given by
\begin{equation}
\bm{\phi}^{l}(x,y) = F^{l}_{fine}(\frac{x}{2^{l}},\frac{y}{2^{l}})
\end{equation}

where $l=0,1,2$ and $F^{l}_{fine}$ is the fine features corresponding to the three scales from corner decoder feature. 
A learnable weight matrix $A_{e}$ can adaptively focus on proposed corners from different scale feature maps.
\begin{equation}
\begin{split}
A_{e}&= \text{softmax}(\Phi(x,y) \cdot W_{e})
\end{split}
\end{equation}
\begin{equation}
\label{PF}
f_{e}=\sum_{l=0}^2A^{l}_{e}\bm{\phi}^{l}(x,y)
\end{equation}
$A_{e}$ is learned from $\Phi(x,y)$ based on a linear layer followed by softmax .
$\Phi\in\mathbb{R}^{3\times128}$ represents the fine features concatenated by $\bm{\phi}(x,y)$.
$f_{e}$ is the corner feature normalized based on weights feature-wise. 

If the learned different channel weights are all 1 for one of the layers, our approach is equivalent to using features only in that layer as our proposal corner feature.
\subsection{Proposal feature enhancement module}
Pairs of corner features are linearly mapped to represent proposal edge feature $f_{v}=W_{v}(f_{e_{1}};f_{e_{2}})$.
It is valid to decode proposal edge feature directly, but the effect is not significant according to Sec. \ref{sec:4.3}.
Our hypothesis is that due to the difficulty of query convergence\cite{DETR,zhou2022centerformer}, the model will be more inclined to obtain image information from global coarse-scale features.
In order to facilitate query convergence and extract fine-grained image features in query, proposal feature enhancement module (PFEM) is designed.

At first, a learnable cosine encoding  is borrowed  from HEAT~\cite{HEAT} as the position embedding $f_{v}^{pos}$.
\begin{equation}
f^{pos}_{v}=  W_{v}^{pos}(\theta_{e_{1}};\theta_{e_{2}})
\end{equation}
where $\theta$ is cosine encoding of the corner coordinates~\cite{VisionTransformer}.
PFEM employs a 6-layers self-attention module of transformer to boost the proposal feature $F_{v}$, whose general formula is:
\begin{equation}
F_{self}=\sum_{m=1}^M W_{m}[\text{softmax}(\frac{QK}{\sqrt{d}})V]
\end{equation}
\begin{equation}
F_{v}^{boost}=Add \& Norm(F_{v},F_{self})
\end{equation}
where $Q=(F_{v}+F_{v}^{pos})W_{Q}$ , $ K=(F_{v}+F_{v}^{pos})W_{K}$,and  $V=F_{v}W_{V}$. 
$F_{self}$ is the output of the self-attention layer.
$F_{v}^{boost}$ is boosted candidate proposal feature. 
$M$ denotes the number of attention self-attention heads and $M=8$ for ours. 
Finally, a feed-forward network (FFN) is used to integrate information and learn fine-grained features on fine-scale feature maps.
\subsection{Edge decoder}

The edge decoder is also 6-layers in total and deformable cross-attention module is adopted to obtain global coarse features from coarse-scale feature maps. 
Each candidate edge boosted $f_{v}^{boost}$  will query the $S$ location features noticed in global coarse features $F_{coarse}$. 
The complete formula for the output of D-Cross Attention $F_{cross}$ in Fig. \ref{fig:method} is expressed as:
\begin{equation}
F_{cross}=\sum_{l=3}^5\sum_{s=1}^{S} A(v)^{boost}_{ls}F_{coarse}^{l}[\bm{p}_{lv}+\Delta_{ls}(f_{v}^{boost})]
\end{equation}
where a linear projection layer is used to learn the corresponding attention position $\Delta(f_{v}^{boost})$ and attention weight $A(v)^{boost}$. 
The initialized attention position $p$ is the center point of the candidate edge to accelerate the query of the $S$ location.
The same FFN layer shared weights by the proposal feature extraction module is employed to learn the feature $F_{edge}$ :
\begin{equation}
F_{edge}=FFN(Add \& Norm(F_{v}^{boost},F_{cross}))
\end{equation}

$F_{edge}$ is eventually used to predict the candidate edges.
We compare the cross-attention module from DETR \cite{DETR} with deformable cross-attention module in our work.
Referring to Sec. \ref{sec:4.3}, deformable cross-attention with $S=4$ is applicable to our work.

\subsection{Loss function}
The final loss function consists of four branches and binary cross entropy (BCE) loss is adopted for all four losses. 
The total loss function is:
\begin{equation}
Loss=\lambda_{1}L_{direct}+\lambda_{2}L_{seg}+L_{boost}+L_{edge}
\end{equation}
1) \mypara{Direction Corner Loss} is used to detect different semantic corners in confidence feature maps. 
2) \mypara{Segmentation Loss} performs semantic segmentation of line segments to assist direction corner detection. 
3) \mypara{Boosting Loss}  supervises proposal features $F_{v}^{boost}$ which facilitate query convergence and extract fine-grained image features.
4) \mypara{Edge Loss} is employed to predict edge based on $F_{edge}$ by the shared FFN. 
$\lambda_{1}$ and $\lambda_{2}$ are the given hyper-parameter to balance the weights.

\section{Experiments}
\input{texts/tables/table_outdoor_ours.tex}

\input{texts/figures/outdoor.tex}
\mypara{Experiment details:}
We implemented the proposed method in PyTorch. 
The learning rate is initialized to 2e-4, which decreases to  1e-5 at 600 epochs with a total of 800 epochs consistent with the previous work \cite{HEAT}.
We train our model with a batch size of 32 for the image size $256\times256$ and a batch size of 12 for the image size $512\times512$ by using the Adam optimizer \cite{Adam}. 
All the experiments are done in dual Tesla V100 GPUs. 
Both loss weight  $\lambda_{1}$ and $\lambda_{2}$ are set as 0.05.
We set the maximum number of training maximum candidate edges to $T=800$ to ensure that all the candidate edges fully participate in the training. 
All candidate edges participate in the prediction during inference.

\mypara{Dataset:}
 Outdoor architecture reconstruction is a building vectorization dataset proposed by Nauata et al. ~\cite{IP}, which leverages to solve the architecture vectorization problems. 
 The input is a satellite RGB image from either Paris, Las Vegas or Atlanta and the output is a planar graph depicting both the internal and external architectural edges in the roof of buildings. 
 The dataset contains 2001 satellite images in total and 1601, 50, 350 for training, validation, testing, respectively. 
 The precision/recall/F-1 scores of the corner/edge/region primitives are the metrics of the dataset. 
 Outdoor architecture reconstruction  is a challenging problem for computer vision because it's not just about learning corner connections similar to floorplan vectorization and further needs to distinguish the target building structure and the edge  segments with other semantics (\eg, shadows, lawns, and non-target structures).
\subsection{Competing methods}
We conduct comparative evaluations against with six competing methods: IP~\cite{IP}, ConvMPN~\cite{ConvMPN}, Exp-cls~\cite{expcls}, HAWP~\cite{HAWP}, LETR~\cite{LETR}, HEAT~\cite{HEAT}. The methods are introduced briefly as follows:

\noindent $\bullet$ {\bf IP}  leverages integer programming algorithm to integrate detected primitives (\eg, corners, edges) and their related information into a planar graph. 
This is the first SOTA algorithm for the current dataset.

\noindent $\bullet$ {\bf ConvMPN} is based on an improved graph neural network in which nodes correspond to building edges in an image. 
This is a single structure detection model for the proposed corners.

\noindent $\bullet$ {\bf Exp-cls}  optimizes the proposed structure from the baseline model ( \eg, IP, Conv-MPN) through iterative exploration and classification. 
The computational framework is general but expensive compared to the end-to-end system.

\noindent $\bullet$ {\bf HAWP} combines Attraction Field Map (AFM) representation method to pre-filter candidate edges.
It predicts edges based on the corner feature map without global features.

\noindent $\bullet$ {\bf LETR} is an end-to-end wireframe parsing algorithm based on transformer framework. 
It inherits the advantages of DETR ~\cite{DETR}  and works end-to-end without edge, connection, region detection, and heuristic guidance.

\noindent $\bullet$ {\bf HEAT} is the current SOTA which designs an end-to-end network system with independent corner detection and edge prediction. 
Our method builds on this to perform the overall learning of corner detection and edge prediction through feature transfer.

Comparisons of all the competing methods are based on publicly available data in original papers. 
In particular, for HAWP and LETR which do not carry official results in the current dataset, we borrow the results from HEAT based on official implementations.
\subsection{Evaluation}
\mypara{Quantitative evaluation:}
Table \ref{tab:outdoor-main} presents the  quantitative evaluation with competing methods.
Our Method CornerFormer in outdoor architecture reconstruction achieved the SOTA. 
Our method focuses more on the attention of all candidate edges. 
In order to realize the direct transfer of boosting corner representation, CorerFormer abandons the filtering edge candidates in HEAT and leverages more resources to realize the two-stage information synchronization. 

When the proportion of memory access is the same, the number of parameters for our model (46.97M) is relatively lower than HEAT (48.94M).
In terms of inference speed, we can exceed the performance from HEAT with three times inference by only one simple inference With the help of the structure information transferred from the corner detection. 
The inference speed is more than twice as fast as the current SOTA, and much faster than heuristic methods, such as IP and Exp-cls.
The default image size of the dataset is 256×256.
In order to compare with HWAP, LETR, and others, we resize the image to 512×512 for train and evaluation. 
Quantitative experiments show that our method has a more significant improvement in region recall, which can recall smaller target regions.

\mypara{Qualitative evaluation:}
For the task of fine-grained structured reconstruction which needs to pay more attention to the detection of "structural corners", ConerFormer shows the extreme advantage in corner proposal. 
After our feature boosting, more fine-grained corners are detected to assist edge prediction. 
Many invalid corners of the graph are filtered out (\eg, invalid structures inside the detected target, as well as non-detected objects) compared to competing methods.

In general, we make a great improvement in maintaining the consistency of the two-stage model and perceiving fine-grained structure, which is particularly important for the vectorization problem of satellite images with complex environments and dense corners.

\input{texts/figures/feature_weight.tex}
\subsection{Ablation studies}
\label{sec:4.3}
\input{texts/tables/table_edge_decoder.tex}

\mypara{Horizontal ablation experiments:}
We investigate the effect of different components designed in our edge prediction model horizontally in Table \ref{tab:boost_feature}. 
We contrast the  representations of different candidate edges, the {\bf endpoints }and the {\bf center-point}, as the candidate edge features. 
Center-point representation is inspired by CenterFormer ~\cite{zhou2022centerformer} with the predicted center-point of the candidate edges.
Experiment results show that 
1) endpoint representations have more explicit semantic information, especially in edges, and can promote the corner detection model to learn fine grained structure;
2) It is sufficient for our task to adopt vanilla convolution for proposal feature extraction compared to deformable convolution with minor improvements at the cost of extra 10 hours' training;
3) We train the same epochs for the cross-attention layer but we found that the deformable cross-attention module is much more effective than the cross-attention module;
4) At the same time, the proposal feature enhanced by self-attention can further extract effective information from the fine features.
It should be noted that $L_{edge}$ is adopted for different types of cross-attention layer alone when the query embedding is position embedding and $L_{boost}$ for self-attention layer.

Figure \ref{fig:weight}  visualizes the feature-wise weights in different scale feature maps, which is learned after the normalization of multi-scale fine features in Eq. \ref{PF}. 
For the convenience of the display, we merge the total 128-dims corner feature into 16-dims. 
It can be seen that the weights learned from different scales are 1:1:2 generally, which proves that the feature in a larger scale is more beneficial for edge prediction.
\input{texts/tables/table_ablation_method.tex}
\input{texts/figures/badcase.tex}
\mypara{Vertical ablation experiments:} We explore each component in CornerFormer through the ablation experiment in Table \ref{tab:ablation}.
Corner feature comes from learnable corner feature extractor and position embedding instead if not.
Experiment results show that the performance of corner feature is significantly by employing $L_{boost}$.  
Direction corner alone is effective for corner detection and performs better with the guidance of $L_{seg}$. 
We analyze that the $L_{seg}$ can further mine the geometric construction in the image to assist direction corner and transfer more semantic information to edge prediction.

\mypara{Limitations:} 
The classical failure cases from ours are shown in Fig.\ref{fig:badcases}. 
Although we have been trying to keep the two-stage model in sync, the model based on the bottom-up architecture itself causes edge prediction to be subject to corner detection. 
Only the proposed corners can be connected as the candidates for edge prediction (\eg, two cases on the left). 
For areas where the image signal is weak, our model cannot be supplemented with logical reasoning like humans (\eg, two cases on the right).

\section{Conclusion}
This paper presents a query-based transformer framework via proposal feature as query embedding for structured reconstruction. 
Our approach not only facilitates the accurate recall of corners but also demonstrates the necessity of learning large-size features for fine-grained structural reconstruction tasks. 
More fine structures in the image can be perceived by proposed fine features. 
We expect this approach to be applied in more areas.
\clearpage

{\small
\bibliographystyle{ieee_fullname}
\bibliography{egbib}
}

\clearpage

\end{document}

%% file: math_definition.tex
\usepackage{amssymb}
\usepackage{amsmath,amsfonts}
\usepackage{amsopn,bm}

\usepackage{nicefrac}       

\usepackage{epsfig}
\usepackage{graphicx}
\usepackage{float}
\usepackage{wrapfig}
\usepackage[ruled,vlined]{algorithm2e}

\usepackage{bm,xspace}
\usepackage{comment}
\usepackage{verbatim}
\usepackage{multirow}
\usepackage{makecell}
\usepackage{array}
\usepackage{balance}
\usepackage{url}
\usepackage{booktabs}
\usepackage{etoolbox,siunitx}
\usepackage{calc}
\usepackage{pifont,hologo}
 
\usepackage{nicefrac}

\usepackage{arydshln}
\usepackage[utf8]{inputenc} 
\usepackage[T1]{fontenc}    
\usepackage{url}            
\usepackage{amsfonts}       
\usepackage{nicefrac}       
\usepackage{microtype}      
\usepackage[american]{babel}
\usepackage{enumitem}
\usepackage{siunitx}

\usepackage{enumitem}

\usepackage[super]{nth}
\usepackage{nicefrac}
\robustify\bfseries

\usepackage{dsfont}

\usepackage{listings}
\usepackage{xcolor}
\definecolor{codegreen}{rgb}{0,0.6,0}
\definecolor{codegray}{rgb}{0.5,0.5,0.5}
\definecolor{codepurple}{rgb}{0.58,0,0.82}
\definecolor{backcolour}{rgb}{1.0,1.0,1.0}

\lstdefinestyle{mystyle}{
    commentstyle=\color{codegreen},
    keywordstyle=\color{magenta},
    numberstyle=\tiny\color{codegray},
    stringstyle=\color{codepurple},
    basicstyle=\ttfamily\footnotesize,
    breakatwhitespace=false,         
    breaklines=true,                 
    captionpos=b,                    
    keepspaces=true,                 
    numbersep=0pt,                  
    showspaces=false,                
    showstringspaces=false,
    showtabs=false,                  
    tabsize=2,
    linewidth=.99\textwidth,
    xleftmargin=0.01cm
}
\usepackage{mathtools}

\newcommand{\cmark}{\text{\ding{51}}}

\newcommand{\ProbOpr}[1]{\mathbb{#1}}

\newcommand{\expect}[2]{%
\ifthenelse{\equal{#2}{}}{\ProbOpr{E}_{#1}}
{\ifthenelse{\equal{#1}{}}{\ProbOpr{E}\left[#2\right]}{\ProbOpr{E}_{#1}\left[#2\right]}}} 
\newcommand{\var}[2]{%
\ifthenelse{\equal{#2}{}}{\ProbOpr{VAR}_{#1}}
{\ifthenelse{\equal{#1}{}}{\ProbOpr{VAR}\left[#2\right]}{\ProbOpr{VAR}_{#1}\left[#2\right]}}} 

\makeatletter
\DeclareRobustCommand\onedot{\futurelet\@let@token\@onedot}
\def\@onedot{\ifx\@let@token.\else.\null\fi\xspace}

\def\eg{\emph{e.g}\onedot}

\def\etal{\emph{et al}\onedot}
\makeatother

\newcommand{\eat}[1]{{}}

\newcommand\mypara[1]{\vspace{1mm}\noindent\textbf{#1}}

%% file: table_definition.tex
\usepackage{tabu}

\definecolor{Gray}{gray}{0.5}

\newlength\savewidth

\makeatletter\renewcommand\paragraph{\@startsection{paragraph}{4}{\z@}
  {.5em \@plus1ex \@minus.2ex}{-.5em}{\normalfont\normalsize\bfseries}}\makeatother

\newcolumntype{x}[1]{>{\centering\arraybackslash}p{#1pt}}
\newcolumntype{y}[1]{>{\raggedright\arraybackslash}p{#1pt}}
\newcolumntype{z}[1]{>{\raggedleft\arraybackslash}p{#1pt}}

\definecolor{Highlight}{HTML}{39b54a}  

%% file: texts/figures/head_img.tex
\begin{figure}[!t]
    \centering
    \includegraphics[width=\linewidth]{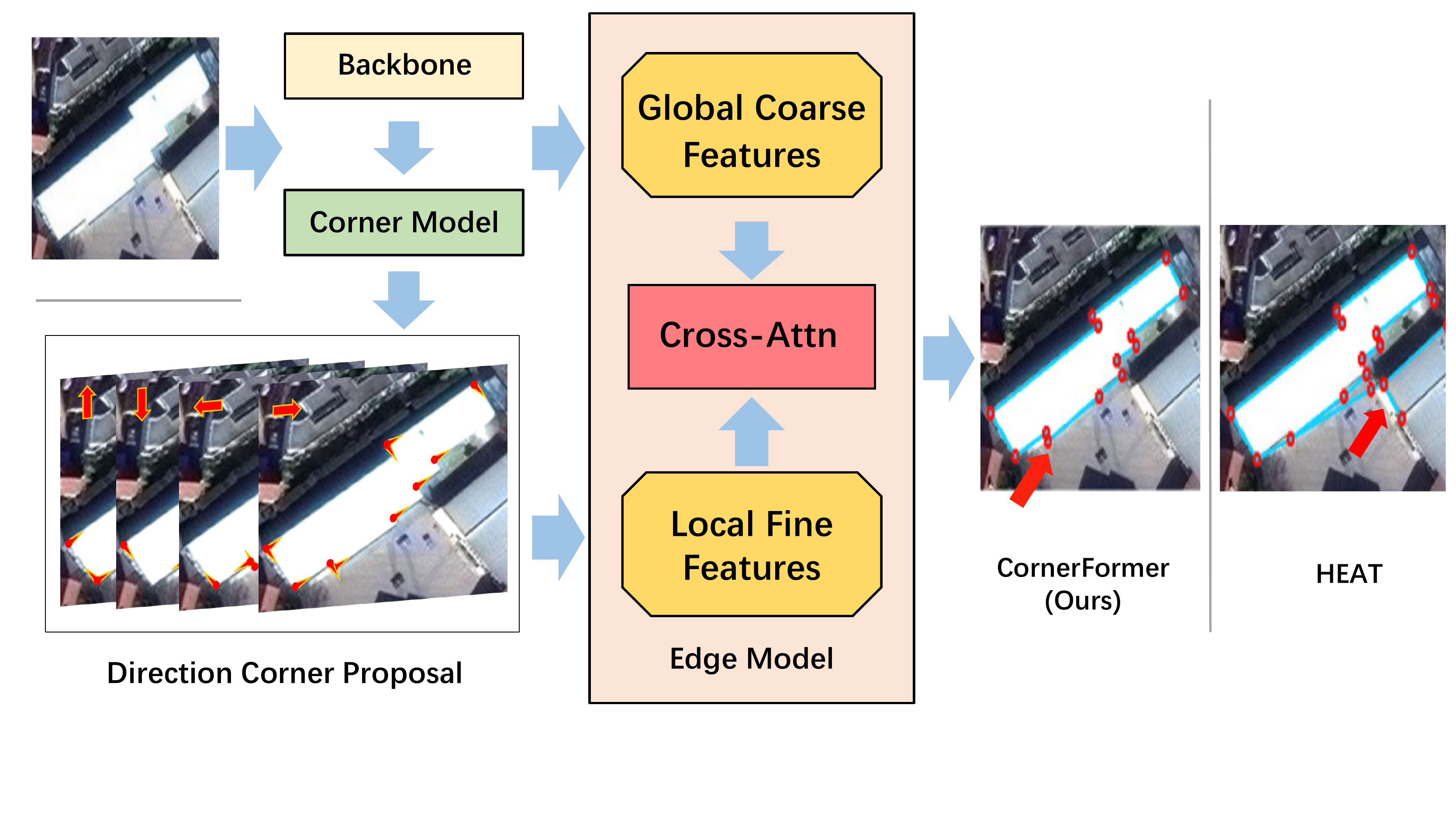}
    \vspace{-2.5em}
     \caption{\textbf{The basic idea of CornerFormer}.
     Red arrows indicate non-target structures and tiny edges.
     CornerFormer perceives both the global information and the local fine grained information.}
    \label{fig:head}
\end{figure}

%% file: texts/figures/method.tex
\begin{figure*}[!ht]
\centering
\includegraphics[width=\linewidth]{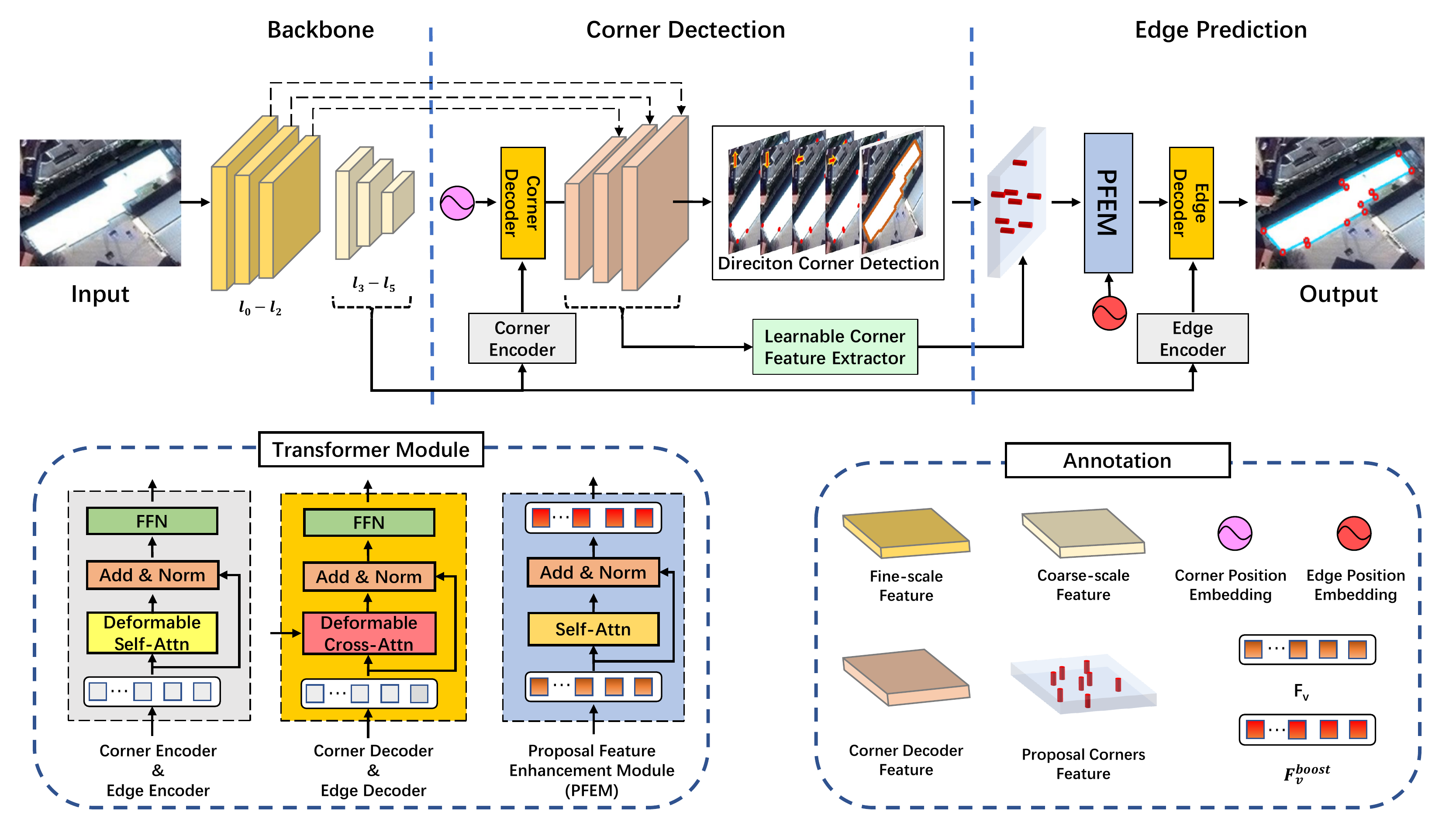}
\vspace{-1.5em}
\caption{\textbf{CornerFormer Architecture.} 
Coarse-scale feature maps $l_{3}-l_{5}$ are assigned to the encoder for extracting \textbf{global coarse features}.
Fine-scale feature maps $l_{0}-l_{2}$ are assigned to the corner decoder for direction corner detection.
The proposed corner features (red cylinders) are extracted by a learnable corner feature extractor.
Proposal feature enhancement module (PFEM) is designed to boost the proposed feature $F_{v}$ and learns \textbf{ local fine features}.
}  
\label{fig:method}
\end{figure*}

%% file: texts/figures/driection_corner.tex
\begin{figure}[!t]
    \centering
    \includegraphics[width=\linewidth]{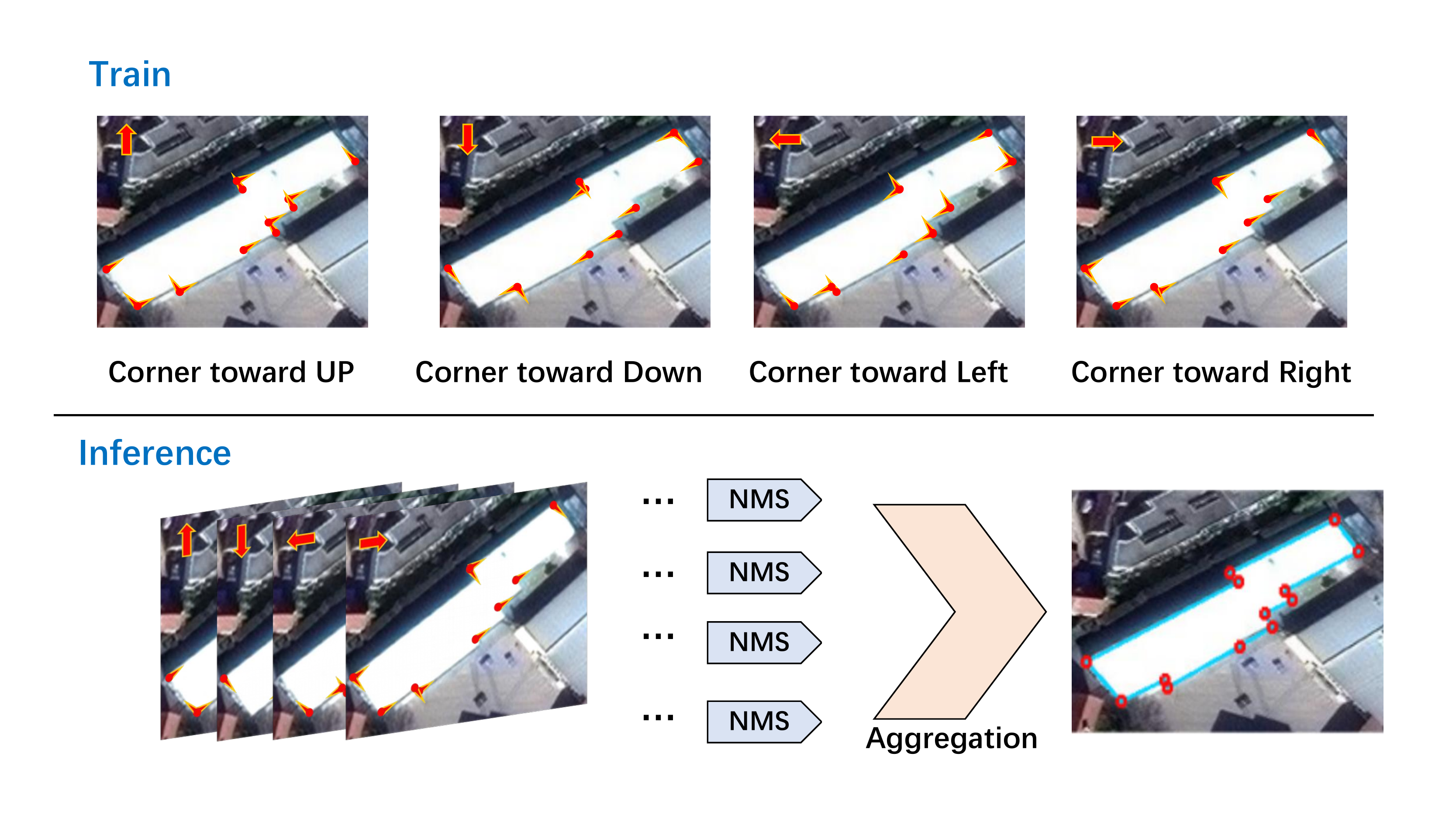}
    \vspace{-2em}
    %
     \caption{\textbf{Direction corner module} for train and inference.
     The direction of the arrow indicates the orientation of the corners.}
    \label{fig:direction_corner}
\end{figure}

%% file: texts/tables/table_outdoor_ours.tex
\begin{table*}[!ht]
    \centering
    \small
    \caption{\textbf{Quantitative evaluations on outdoor architecture reconstruction.} Size: the size of the input image.
    End-to-end: a complete end-to-end neural system for both corner proposal and edge prediction.
    The colors \first{cyan} and \second{orange} mark the top-two results with different image sizes. 
    }
    {
        \tabcolsep 5pt
	    \begin{tabu}{@{\;} lcc @{\quad\;\;\;\;}  ccccccccc@{\;}}
	        \addlinespace
	        \toprule
	        \multicolumn{3}{l}{Evaluation Type} &  \multicolumn{3}{c}{Corner} &  \multicolumn{3}{c}{Edge} &  \multicolumn{3}{c}{Region} \\
	        \cmidrule(lr){4-6} \cmidrule(lr){7-9} \cmidrule(lr){10-12}
	        {Method} & Size & End-to-end &  Prec & Recall & F-1 & Prec & Recall & F-1 & Prec & Recall & F-1  \\
	        \midrule
	        IP~\cite{IP} & 256 & - &  - & - & 74.5 & - & - & 53.1 & - & - & 55.7 \\
	        ConvMPN~\cite{ConvMPN} & 256 &  - &  78.0 & 79.7 & 78.8 & 57.0 & 59.7 & 58.1 & 52.4 & 56.5 & 54.4  \\ 
                Exp-Cls~\cite{expcls} & 256 &  - &  \second{92.2} & 75.9 & 83.2 & 75.4 & 60.4 & 67.1 & 74.9 & 54.7 & 63.5 \\
	        HAWP~\cite{HAWP} & 256 &  \cmark &  90.9 & 81.2 & 85.7 & 76.6 & 68.1 & 72.1 & 74.1 & 55.4 & 63.4 \\
	        LETR~\cite{LETR} & 256 &  \cmark & 87.8 & 74.8 & 80.8 & 59.7 & 58.6 & 59.1 & 68.3 & 48.7 & 56.8 \\
	        HEAT~\cite{HEAT} & 256 & \cmark & 91.7 & \second{83.0} & \second{87.1} & \second{80.6} & \second{72.3} & \second{76.2} & \second{76.4} & \second{65.6} & \second{70.6} \\
            \ourmethodtable & 256 &  \cmark & \first{94.1} & \first{84.5} & \first{89.0} & \first{83.8} & \first{75.1} & \first{79.2} & \first{77.8} & \first{66.9} & \first{71.9}  \\
	        \midrule
	        HAWP~\cite{HAWP} & 512 &  \cmark & 90.6 & 83.7& 87.0& 78.8 & 72.0 & 75.2 & 77.5 & 57.8 & 66.2 \\
	        LETR~\cite{LETR} & 512 &  \cmark & 90.3 & 79.7 & 84.7 & 64.0 & 71.6 & 67.6 & 77.1 & 62.6 & 69.1 \\
            HEAT~\cite{HEAT} & 512 &  \cmark & \second{90.7} & \second{86.7} & \second{88.7} & \second{82.2} & \second{77.4} & \second{79.7} & \first{79.6} & \second{69.0} & \second{73.9}  \\
	        \ourmethodtable & 512 &  \cmark & \first{92.6} & \first{87.8} & \first{90.2} & \first{83.3} & \first{79.4} & \first{81.3} & \second{79.4} & \first{72.8} & \first{76.0}  \\
    	    \bottomrule
	    \end{tabu}
	}
    \label{tab:outdoor-main}
\end{table*}

%% file: texts/figures/outdoor.tex
\begin{figure*}[!t]
    \centering
    \includegraphics[width=0.97\textwidth,height=\textwidth]{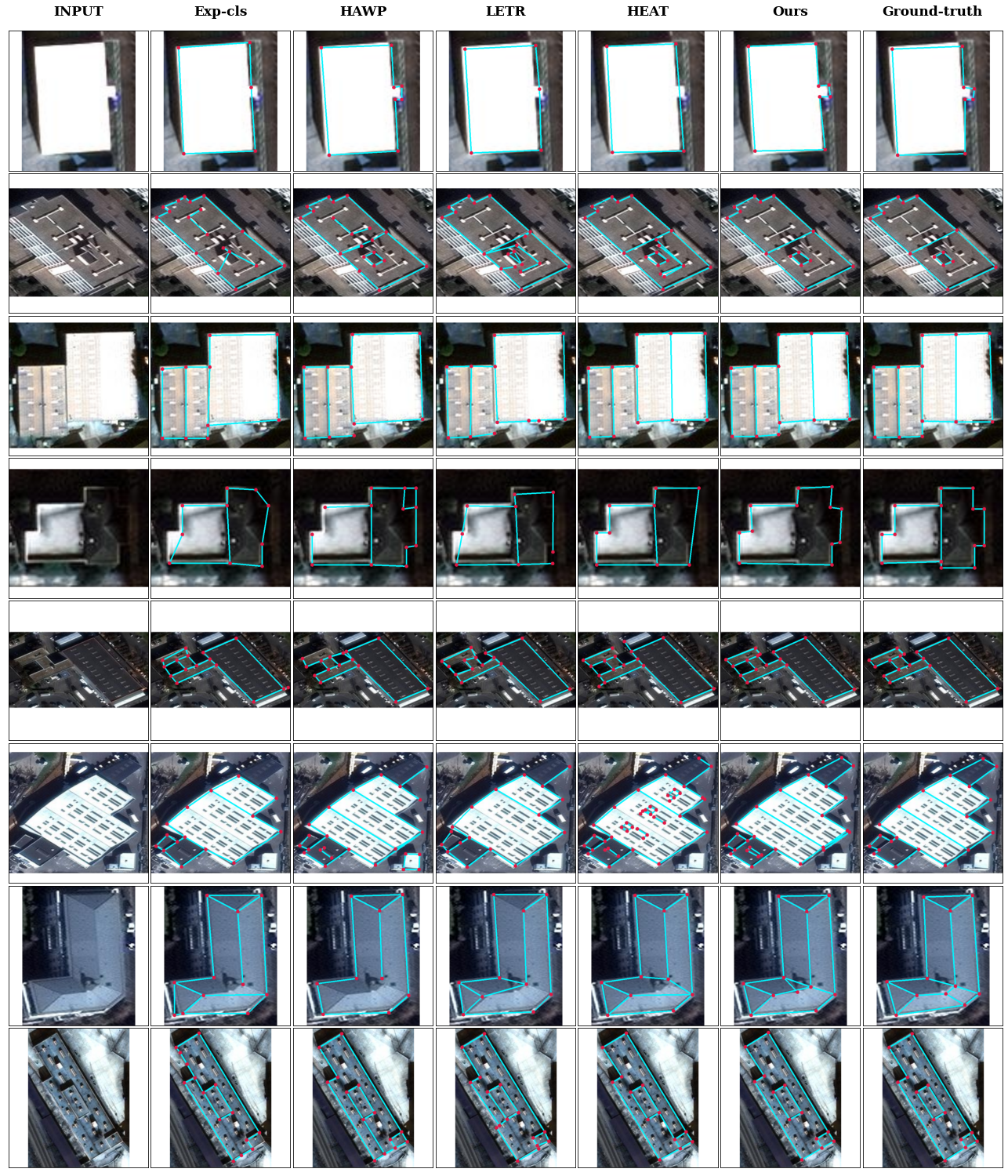}
    \vspace{-1em}
    \caption{\textbf{Qualitative evaluations on outdoor architecture reconstruction} with image size 256$\times$256. Our models provide better perception of fine structures, especially corners. }
\label{fig:outdoor-qualitative}
\end{figure*}

%% file: texts/figures/feature_weight.tex
\begin{figure}[!t]
    \centering
    \includegraphics[width=\linewidth,height=0.25\linewidth]{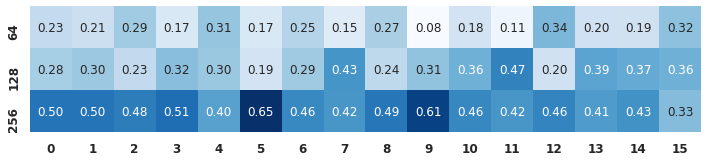}
    \vspace{-2em}
     \caption{\textbf{The feature-wise weights from fine features for proposal corner feature.}
     We analyze the weights from our model with the input is 256× 256.
     The darker the color, the larger the proportion of the corresponding scale feature.}
     
    \label{fig:weight}
\end{figure}
     

%% file: texts/tables/table_edge_decoder.tex
\begin{table}[!ht]
    \setlength{\tabcolsep}{2pt}
    \centering
    \small
    \caption{\textbf{The ablation of different components designed in our edge prediction.}
    The evaluation metric is the scores of corner/edge/region F-1. 
    CER denotes the candidate edge representation which applies endpoints (EP) or center-point (CP). 
    CN and DCN denote vanilla convolution and deformable convolution in corner feature extractor.
    SA, CA, and DCA denote the self-attention layer form PFEM, cross-attention layer, and deformable cross-attention layer from edge decoder.}
    {
        \begin{tabular}{cc|cc|ccc|ccc}
	        \addlinespace
	        \toprule
	        \multicolumn{2}{c}{CER} & \multicolumn{2}{c}{Convolution} & \multicolumn{3}{c}{Transformer} & \multicolumn{3}{c}{Evaluation}\\
	        \cmidrule(lr){1-2} \cmidrule(lr){3-4} \cmidrule(lr){5-7}\cmidrule(lr){8-10}
	      EP & CP & CN & DCN & SA & CA & DCA & Corner & Edge & Region \\
            \hline
            \cmark &  & \cmark &  &  & \cmark &  & 86.9 & 74.9 & 58.3 \\
            \cmark &  & \cmark & &  & & \cmark & 87.0 & 75.6 & 66.2 \\
            \hline
             & \cmark  & \cmark &  & \cmark &  &  & 87.2 & 75.3 & 67.0 \\
            \cmark &  & \cmark &  & \cmark &  &  & 87.5 & 76.2 & 69.5 \\
            \hline
            \cmark &  & \cmark &  & \cmark & \cmark &  & 87.6 & 77.0 & 67.9 \\
            \cmark &  & \cmark &  & \cmark &  & \cmark & \textbf{87.7} & \textbf{76.9} & \textbf{70.8} \\
            \cmark &  &  & \cmark & \cmark &  & \cmark & 87.9 & 77.4 & 70.7 \\
    	    \bottomrule
	    \end{tabular}
	}
    \label{tab:boost_feature}
\end{table}

%% file: texts/tables/table_ablation_method.tex
\begin{table}[!t]
    \centering
    \small
    \caption{\textbf{The ablation of each component in CornerFormer.} 
    The evaluation metric is the scores of corner/edge/region F-1. $L_{boost}$,$L_{direct}$,$L_{seg}$ are the loss functions adopted by our method respectively. 
    Corner Feature with $L_{boost}$ denotes the selected method for edge prediction (The bold one in Table \ref{tab:boost_feature}).
    \vspace{-1em}
    } 
    {
        \tabcolsep 3pt
	    \begin{tabu}{@{\;} cccc | ccc@{\;}}
	        \addlinespace
	        \toprule
	        Corner Feature  &  $L_{boost}$ & $L_{direct} $& $L_{seg} $ & Corner & Edge & Region \\
            \midrule
            & & & & 86.9 & 75.1 & 64.3\\
            & &\cmark & & 87.7 & 76.7 & 66.5\\
            \cmark& & & & 87.6 & 76.4 & 68.3\\
            \midrule
            \cmark &\cmark & &  &87.7  & 76.9 & 70.8\\
            \cmark& \cmark& & \cmark& 87.8 & 77.5 & 70.2 \\
            \cmark& \cmark& \cmark& & 88.2 & 78.0 & 71.4\\
            \cmark& \cmark& \cmark& \cmark& 89.0 & 79.2 & 71.9\\
        \bottomrule
	\end{tabu}
	}
    \label{tab:ablation}
\end{table}

%% file: texts/figures/badcase.tex
\begin{figure}[!ht]
    \centering
    \includegraphics[width=\linewidth]{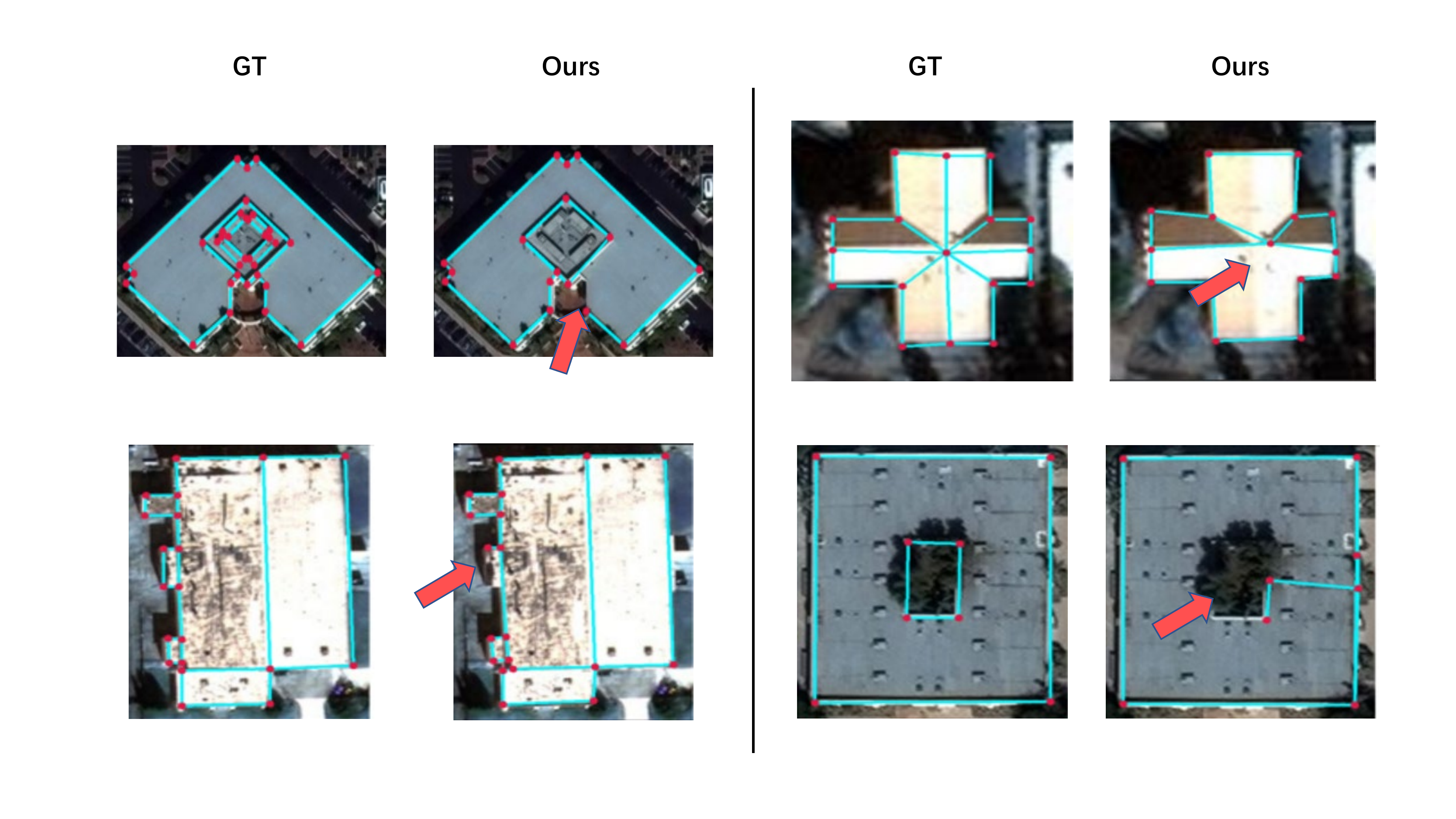}
    \vspace{-2em}
     \caption{\textbf{The failure cases.} 
     The two cases on the left are due to the lack of corners proposed by corner detection and the other two on the right are due to the inconspicuous image signal in the outdoor  structure.}
     
    \label{fig:badcases}
\end{figure}
     